\documentclass[letterpaper]{article} 
\usepackage{aaai2027}  
\usepackage[hyphens]{url}  
\usepackage{graphicx} 
\usepackage{natbib}  
\usepackage{caption} 
\usepackage{algorithm}
\usepackage{algorithmic}
\usepackage{multirow}
\usepackage{newfloat}
\usepackage{listings}
\DeclareCaptionStyle{ruled}{labelfont=normalfont,labelsep=colon,strut=off} 
\floatstyle{ruled}
\newfloat{listing}{tb}{lst}{}
\floatname{listing}{Listing}

\usepackage{booktabs}
\usepackage{amsmath}
\usepackage{amssymb}
\usepackage{array}
\usepackage{tabularx}
\usepackage{xcolor}
\nocopyright 

\title{StairMaster: Learning to Conquer Risky Hollow Stairs \\
for Agile Quadrupedal Robots}
\author{
    Xincheng Tang$^{1}$, Youhan Xie$^{1}$, Zhengjie Shu$^{1}$, Wanyu Li$^{1}$, Lai Jiang$^{1}$, Wenkang Hu$^{1}$, Yitong Li and Ruigang Yang$^{1,\corresponding}$
}
\affiliations{
    \textsuperscript{\rm 1}Shanghai Jiao Tong University\\

}

\begin{document}

\maketitle

\begin{abstract}
Climbing hollow stairs remains a challenging problem for quadruped robots due to the high risk of leg trapping, severe depth sparsity, and high-frequency depth-sensing noise. In this paper, we propose StairMaster, a three-stage reinforcement learning framework for stable locomotion on such extreme discontinuous terrains.  Our architecture integrates a Cross-Attention mechanism to extract structural features from noisy depth data, alongside a Spatial-aware Recurrent Unit (SRU) that maintains robust spatio-temporal memory to mitigate perception blind spots. To bridge the sim-to-real gap in depth perception, we propose a high-fidelity sim-to-real depth sensor modeling pipeline that faithfully replicates real-world sensor artifacts. Additionally, we employ a 3D waypoint-guided active perception reward for proactive sensing, alongside hollow gap kinematic and stair edge penalties to ensure precise foothold placement. We successfully deployed StairMaster on a Unitree Go2 robot, demonstrating its ability to conquer hollow stairs with an unprecedented incline of up to 55$^\circ$ through zero-shot transfer. To the best of our knowledge, this is the first RL-based policy to achieve such steep hollow stair climbing in real-world environments. 
\end{abstract}


\section{INTRODUCTION}

In nature, quadrupedal animals exhibit a remarkable ability to stably navigate highly complex terrains. Inspired by this biological agility, deep reinforcement learning (DRL) has enabled quadruped robots to traverse unstructured environments by fusing visual perception with proprioception \cite{rudin2022learning,kumar2021rma,peng2020learning,wu2023learning}. In industrial facilities such as power plants and construction sites, hollow stairs are widely used to reduce structural weight. While quadruped robots have substantial application potential across diverse industrial scenarios, climbing hollow stairs remains a challenging problem in legged robotics.

Unlike standard solid stairs, hollow stairs feature large open gaps between treads, introducing three critical challenges. First, the absence of vertical risers creates a severe risk of leg trapping; even a minuscule foothold error can cause the robot's leg to plunge into the void, leading to catastrophic hardware damage. Second, these stairs are often constructed from reflective or grating materials, inducing severe depth sparsity and causing depth cameras to suffer from massive pixel dropouts and extreme noise. Finally, due to the limited field of view of forward-facing cameras, the narrow treads completely disappear as they pass beneath the robot, forcing it to execute extremely precise hind-leg placements under total visual occlusion.

\begin{figure}[t] 
   \centering
   \includegraphics[width=0.95\columnwidth]{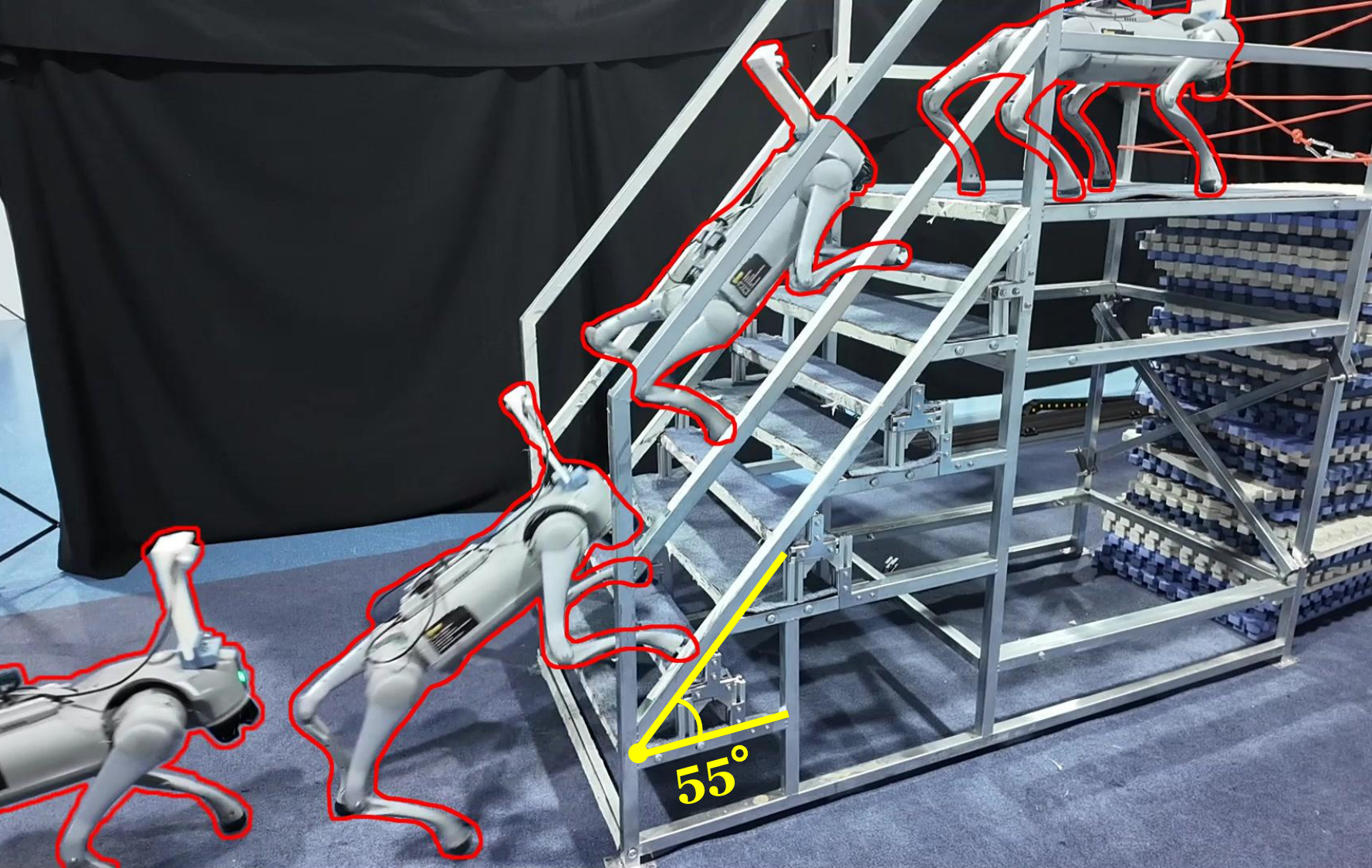}
   \caption{StairMaster enables a quadrupedal robot to climb real-world 55$^\circ$ hollow stairs.}
   \label{fig:titu}
\end{figure}

Existing visuo-motor control frameworks struggle significantly when confronted with the compounding challenges of real-world deployments \cite{zhuang2023robot,yang2025agile,cheng2024extreme}. First, they must overcome severe depth sensor noise, which is further exacerbated by rapid trunk rotations and impact-induced camera oscillations during stair climbing \cite{li2025kivi, zhang2025renet,rudin2025parkour,zhu2026hiking}. Second, they suffer from critical partial observability; due to the forward-facing camera, the robot completely loses sight of the terrain beneath its feet and the hollow gaps as the treads pass out of view \cite{rudin2025parkour,zhu2026hiking}. Navigating such prolonged blind spots necessitates robust, simultaneous temporal and spatial memory. While traditional memoryless policies or simple frame-stacking fail to maintain accurate 3D spatial representations, standard RNNs also exhibit limitations in spatial awareness. They primarily encode one-dimensional temporal dependencies and lack the capability to implicitly align global spatial topology under the robot's ego-motion \cite{yang2025spatially}. Furthermore, conventional reward designs and 2D waypoint-guided frameworks fail to encourage active visual perception. They cannot proactively guide the robot to adjust its pitch angle to observe upcoming stairs, which is a critical capability for conquering exceptionally steep inclines.

To overcome the aforementioned challenges, we propose StairMaster, an end-to-end three-stage reinforcement learning framework designed to conquer high-risk hollow stairs. To address visual blind spots and perception degradation, we introduce a novel visuospatial encoder that combines a Cross-Attention mechanism to extract sparse structural features with a Spatial-Aware LSTM \cite{yang2025spatially}.  A high-fidelity depth sensor modeling pipeline is proposed to bridge the depth sim-to-real gap. Furthermore, we introduce three customized reward functions tailored for hollow stairs, including a 3D waypoint-guided active perception reward for proactive sensing, alongside hollow gap kinematic and stair edge penalties to ensure precise foothold placement. We deployed StairMaster on a Unitree Go2 robot and validated across diverse real-world hollow stairs, achieving robust performance through zero-shot transfer. Our technical contributions are:
\begin{itemize}
    \item We present StairMaster, a three-stage reinforcement learning framework specifically designed for high-risk hollow stairs. Extensive simulations and real-world experiments on a Unitree Go2 robot demonstrate that StairMaster achieves robust, zero-shot sim-to-real transfer. More importantly, to the best of our knowledge, this is the first RL-based strategy that enables a quadruped robot to conquer hollow stairs with an unprecedented incline of up to 55$^\circ$.
    \item We introduce a visuospatial encoder integrating a Cross-Attention mechanism with a Spatial-Aware LSTM to extract structural features and maintain robust spatio-temporal memory, overcoming the limited field of view from sensors. 
    \item We carefully design a depth sensor simulation module to account for sensing artifacts caused by abrupt motion and challenging environments, which include thin structures, hollow spaces, and reflective metallic surfaces. It greatly reduces the sim-to-real domain gap, enabling zero-shot transfer. 
\end{itemize}

\section{RELATED WORK}

\begin{figure*}[t] 
   \centering
   \includegraphics[width=0.9\textwidth]{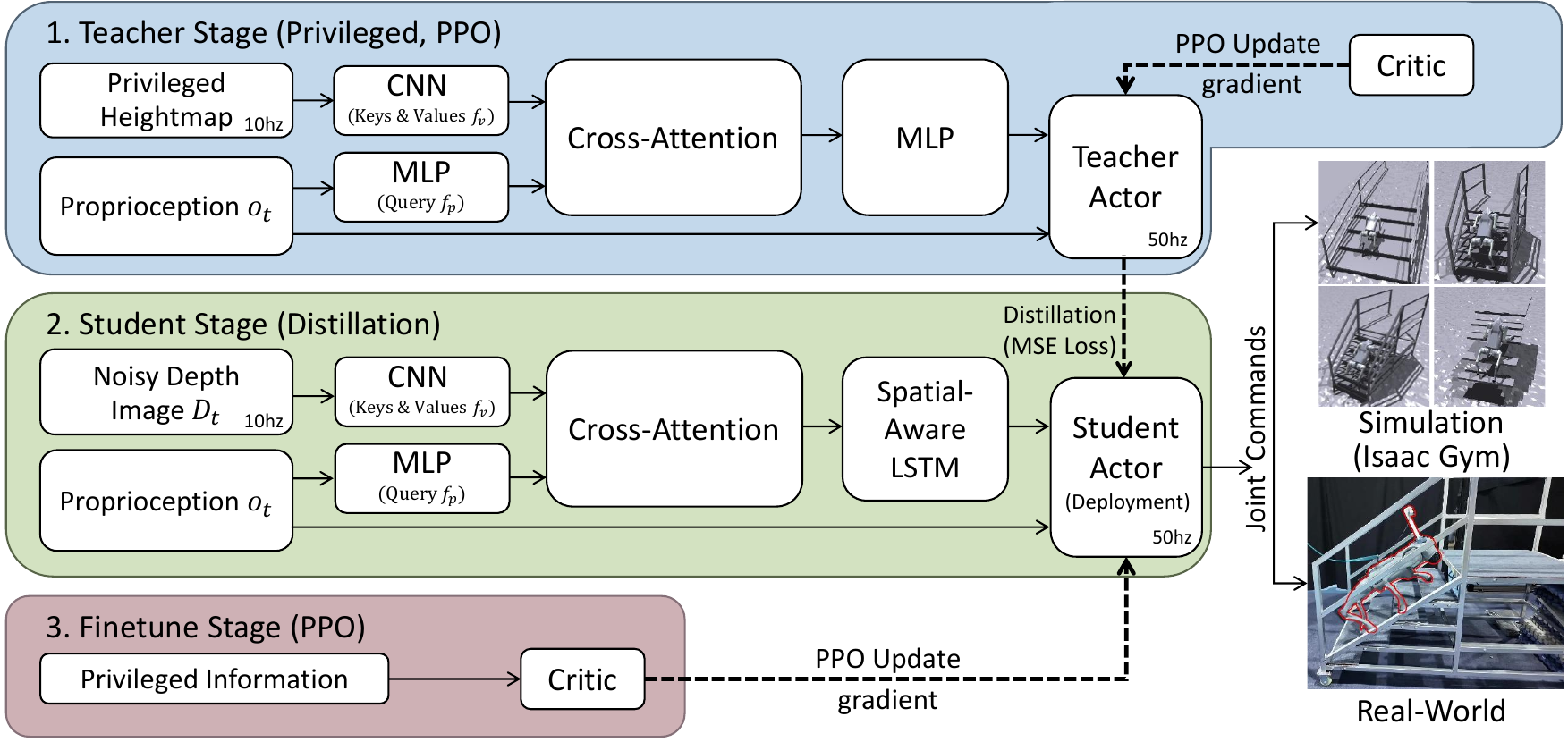}
   \caption{Overview of StairMaster training framework.}
   \label{fig:full_structure}
\end{figure*}

\subsection{Proprioceptive Locomotion}
Robust locomotion using purely proprioceptive sensing has laid the foundation for legged robot control. Traditionally, model-based optimization methods enabled basic walking but often struggled with the unpredictability of complex terrains \cite{bledt2018cheetah}. To overcome this, massive parallel DRL in simulation has allowed robots to learn fundamental walking skills within minutes \cite{rudin2022learning}. Building on this, researchers have introduced various frameworks to estimate latent environmental properties without exteroception, such as motor adaptation \cite{kumar2021rma, margolis2024rapid,ji2022concurrent}, implicit terrain imagination \cite{nahrendra2023dreamwaq}, collision detection \cite{zhu2025robust} and hybrid internal models \cite{long2023hybrid}. Moreover, a specific hook was designed to handle climb ladders \cite{vogel2025robust}. While these proprioceptive-only policies achieve fairly robustness, which allows them to traverse uneven terrains and standard solid stairs \cite{lee2020learning,fu2023deep}, they remain inherently blind. This reactive nature becomes unreliable on sparse and discontinuous
structures such as hollow stairs, where the lack of predictive terrain awareness prevents early adaptation of the trunk orientation and leg configuration, increasing the risk of stepping into open gaps.

\subsection{Vision-guided Locomotion}
The integration of exteroception, such as depth vision and LiDAR, has significantly pushed the boundaries of quadrupedal agility. Early LiDAR-based methods using elevation maps achieved robust locomotion in complex and confined spaces \cite{hoeller2024anymal,he2025attention,zhang2026ame}, yet they often suffer from high computational overhead, latency, and heavy reliance on precise state estimation. To address these issues, vision-based end-to-end frameworks have gained prominence, enabling robots to perform challenging tasks such as solid stair climbing and gap crossing \cite{zhuang2023robot,cheng2024extreme,yang2021learning,agarwal2023legged}. Yang et al. \cite{yang2025agile} further integrated RL with motion tracking, but the policy is limited by a fixed gait. While methods like Extreme Parkour \cite{cheng2024extreme} employ a two-stage teacher-student distillation framework, their resilience to real-world sensor noise remains limited. Conversely, single-stage frameworks such as PIE \cite{luo2024pie} and WMP \cite{lai2025world} eliminate the need for privileged teachers but often face slow convergence and prolonged training cycles. More recently, PLANC \cite{dai2026walk} introduced a three-stage training pipeline to address these perception-action challenges; however, its application is primarily optimized for humanoid robots and does not account for the specific geometric hazards of quadrupedal hollow-stair climbing.

Despite these successes, current vision-guided policies are primarily optimized for solid stairs and fail on hollow ones due to three key deficiencies: critical blind spots beneath the torso caused by front-facing cameras, a lack of spatio-temporal memory to maintain the spatial connectivity of sparse stair treads over time, and high sensitivity to depth noise during violent oscillations. Furthermore, existing methods lack active visual perception to capture anticipated upcoming structures, nor do they explicitly address the severe scarcity of valid footholds caused by the large vertical and horizontal gaps in such discontinuous terrains. 

\section{METHOD}

The proposed StairMaster framework is an end-to-end multi-stage learning-based framework that derives desired joint angle commands directly from raw depth images and onboard proprioception. Our framework introduces a specialized three-stage pipeline tailored to the unique challenges of extreme hollow stairs. This approach enhances the robot’s robustness by integrating high-fidelity depth noise modeling for realistic perception, and building spatio-temporal memory to bridge observation gaps. An overview of the framework and its technical details are provided below.

\subsection{Overview}

As shown in Figure~\ref{fig:full_structure}, the StairMaster framework is structured around a three-stage training pipeline designed to enable stable quadrupedal locomotion on hollow stairs.

\subsubsection{Stage 1: Privileged Teacher Policy Training}

In the initial stage, we train a teacher policy which has access to proprioception and privileged heightmaps through the Proximal Policy Optimization (PPO) algorithm. Leveraging our customized reward functions, this stage yields a highly capable the teacher policy that serves as an expert for student-policy distillation.

Throughout our framework, the proprioceptive state of the robot is explicitly defined as $o_t = [\omega_t, g_t, c_t, \theta_t, \dot{\theta}_t, a_{t-1}]^T$. This observation vector encompasses the base angular velocity $\omega_t \in \mathbb{R}^3$, the projected gravity vector in the body frame $g_t \in \mathbb{R}^3$, the velocity command $c_t \in \mathbb{R}^3$, the joint positions $\theta_t \in \mathbb{R}^{12}$, the joint velocities $\dot{\theta}_t \in \mathbb{R}^{12}$, and the previous actions $a_{t-1} \in \mathbb{R}^{12}$.

\subsubsection{Stage 2: Student Policy Distillation}

In real-world deployment, the robot can only rely on noisy egocentric depth images and proprioceptive states. To transfer the expert behaviors to the real robot, this second stage trains a student policy to imitate the teacher's actions via a distillation process using Mean Squared Error (MSE) loss. To handle the depth sparsity of hollow stairs, the student policy uses Cross-Attention to fuse depth and proprioceptive features, an SRU to retain temporal terrain context, and a depth noise modeling pipeline to improve sim-to-real transfer.

\subsubsection{Stage 3: Fine-Tuning}

Distillation alone often results in a performance gap. To maximize the robot's agility and robustness on challenging hollow stairs, the third stage fine-tunes the distilled student policy using the PPO algorithm. This final RL-based fine-tuning stage allows the student network to interact directly with the environment, exploring and correcting the sub-optimal behaviors from the distillation, ultimately outputting the final target joint actions for zero-shot real-world deployment.

\subsection{Visuospatial Encoder Architecture}

Relying on a single front-mounted depth camera inherently introduces severe visual blind spots during locomotion. Consequently, critical terrain features, such as narrow stair treads, frequently disappear from the sensor's field of view prior to the hind legs engaging with them. To mitigate the challenges posed by this partial observability, we introduce a visuospatial encoder designed to seamlessly fuse multimodal exteroceptive and proprioceptive inputs. This module not only infers immediate terrain geometry but also actively maintains a robust, spatio-temporal memory of the traversed environment to guide stable stepping behaviors in completely occluded regions.

\subsubsection{Multimodal Feature Extraction and Cross-Attention}
At each timestep, the encoder processes the most recent depth frame $D_t$ through a CNN to extract a dense visual feature map $f_v$. Concurrently, the robot's current proprioceptive state $o_t$ is embedded via a MLP into a high-dimensional latent vector $f_p$. Recognizing that simple feature concatenation often incorporates a significant amount of task-irrelevant depth pixels, we achieve multimodal integration through a specialized Multi-head Cross-Attention mechanism. By formulating the proprioceptive embedding ${f}_p$ as the query, and the visual feature map ${f}_v$ as the corresponding keys and values, the model computes state-dependent attention weights. This architectural design enables the encoder to dynamically focus on task-critical geometric structures such as sparse stair edges that are most relevant to the robot's current kinematic posture, ultimately yielding a highly distilled, state-aware latent representation.

\subsubsection{Spatio-Temporal Memory with Spatial-Aware LSTM}
Due to the camera's forward-facing configuration, the Cross-Attention mechanism only extracts instantaneous local features, lacking the spatio-temporal memory required to track stair treads into visual blind spots. To construct a comprehensive environmental representation, we process the fused features through a Spatial-Aware LSTM (SRU). Unlike standard RNNs that struggle with spatial registration, this module integrates spatial snapshots over time without requiring explicit ego-motion inputs \cite{yang2025spatially}. 

Specifically, it introduces a learnable spatial transformation gate, $s_t$, to implicitly align historical memory features with the current perspective. Given the current fused feature ${f}_t$, $s_t$ is computed as:
$$ s_t = \sigma({W}_s {f}_t + {b}_s), \eqno{(1)}$$
where $\sigma$ is the sigmoid function, and ${W}_s, {b}_s$ are learnable parameters. This spatial gate is then applied to the previous hidden state $h_{t-1}$ and cell state $c_{t-1}$ via the Hadamard product ($\ast$). The complete recurrence is formulated as:
$$ h_t, c_t = \text{LSTM}({f}_t, s_t \ast h_{t-1}, s_t \ast c_{t-1}).\eqno{(2)}$$
By continuously registering frame-by-frame representations, this mechanism builds a robust full-scale spatio-temporal memory buffer. It empowers the downstream actor network to retain task-relevant terrain information after the corresponding treads leave the camera field of view and generate reliable joint commands.

\subsection{High-Fidelity Sim-to-Real Depth Noise Modeling}

\begin{figure*}[t] 
   \centering
   \includegraphics[width=\textwidth]{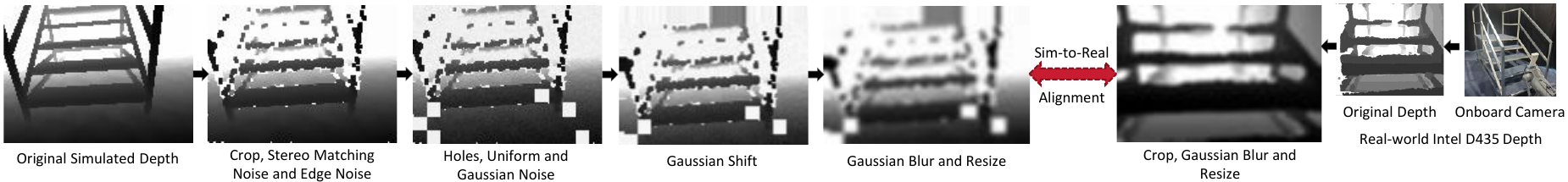}
   \caption{Pipeline of high-fidelity sim-to-real depth noise modeling.}
   \label{fig:Depth}
\end{figure*}

Pristine depth maps rendered in physics simulators fundamentally differ from real-world depth sensor outputs, which are notoriously plagued by scattering, absorption, and motion-induced artifacts. To successfully deploy the vision-guided student policy on physical hardware, bridging this severe visual sim-to-real gap is imperative. As illustrated in Figure~\ref{fig:Depth}, we implement a comprehensive depth noise modeling pipeline during the training of the visuospatial Encoder to bridge the sim-to-real gap.

\subsubsection{Spatial and Sensor Artifact Modeling}
Real-world depth sensors frequently fail on reflective surfaces or sharp geometric transitions, which are highly prevalent in hollow stair environments. We corrupt the ideal simulated depth images with a mixture of spatial noise models:

\begin{itemize}
\item Gaussian and Uniform Noise: We apply a combination of additive Gaussian and uniform noise to simulate the baseline measurement uncertainty and thermal noise inherent in IR-based depth hardware.
\item Hole Noise: Hole noise (random pixel dropout) is utilized to simulate regions where infrared rays fail to return such as the reflective surfaces.
\item Edge Noise: Specifically for the thin, discontinuous geometry of open-riser stairs, we apply localized edge noise to simulate the bleeding effect and scattering typical of depth discontinuities.
\end{itemize}

\subsubsection{Disparity and Dynamic Vibration Noise}
Beyond static artifacts, we explicitly model dynamic noise arising from impulsive foot–ground contacts and impact-induced vibrations during locomotion. 
\begin{itemize}
\item Stereo Matching Noise: We simulate quantization and matching errors typical of binocular depth estimation. This noise is treated as a dynamic artifact because rapid perspective changes and motion blur during climbing lead to transient matching ambiguities on the sparse and narrow stair treads.
\item Gaussian Shift: Aggressive impacts induce severe vibrations in the head-mounted camera. We introduce a Gaussian shift noise that randomly translates depth pixels spatially across consecutive frames. This forces the Cross-Attention and Spatial-Aware LSTM modules to learn vibration-invariant geometric representations rather than overfitting to unstable pixel-wise locations.
\end{itemize}

\subsubsection{Depth Preprocessing}
We apply a unified preprocessing pipeline to both simulated and real depth before they are fed into the neural network. Specifically, the depth images undergo size cropping, depth value clipping, spatial resize, and Gaussian blur. This standardized alignment ensures consistent inputs for a robust transfer to physical hardware.

\subsection{Customized Reward Design for Hollow Stairs}

Standard rewards struggle on steep hollow stairs due to severe foot-trapping risks and poor visual awareness \cite{long2023hybrid,cheng2024extreme}. To address this, we introduce three customized reward terms in addition to the standard reward set during the Stage 1 and Stage 3 training phases to safely navigate these geometric hazards.

\subsubsection{3D Waypoint-Guided Active Perception Reward ($r_{\text{pitch}}$)}

\begin{figure}[t] 
   \centering
   \includegraphics[width=1\columnwidth]{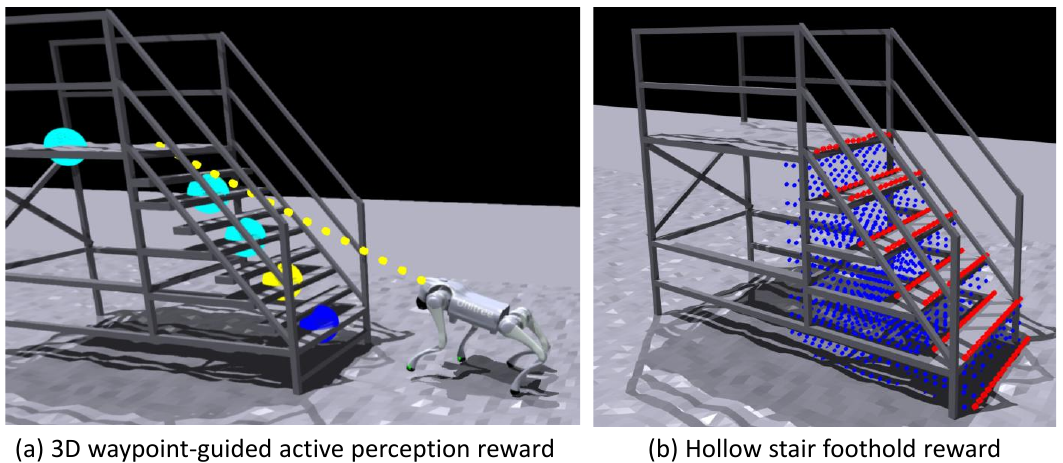}
   \caption{Designed reward for hollow stairs.}
   \label{fig:foothold}
\end{figure}

Previous waypoint-based locomotion frameworks predominantly utilize 2D waypoints to guide the robot's yaw angle for directional heading \cite{cheng2024extreme}. However, such 2D guidance remains insufficient for the extreme geometry of steep hollow stairs. As illustrated by the yellow dashed line in Figure~\ref{fig:foothold}(a), we introduce a 3D waypoint tracking mechanism that targets the center of the second upcoming stair tread ${p}_{\text{target}}$. By calculating the relative vector between the robot's base and this elevated forward waypoint, we directly supervise both the yaw and, crucially, the pitch angle ($\theta_{\text{pitch}}$).

This 3D look-ahead formulation serves a dual purpose:\begin{itemize}\item Kinematic Optimization: Adjusting the trunk pitch reorients the robot relative to the steep stair
surface, providing a more favorable leg workspace for swing-foot clearance and foothold placement during ascent.\item Active Perception: This mechanism acts as an implicit active sensing strategy. By guiding the robot to look up toward the distant waypoint, the onboard egocentric camera captures the structural features of upcoming treads much earlier than a horizontal baseline. This foresight is critical for the visuospatial encoder to populate its spatial memory before the treads enter the camera's near-field blind spots.\end{itemize}

To prevent the robot from pitching its base upwards prematurely while still traversing flat ground, we introduce a distance-based activation threshold $d_{\text{th}}$. The active pitch guidance is only triggered when the Euclidean distance $d$ between the robot and the upcoming waypoint falls below this threshold. We formulate this reward using a piecewise function to penalize the deviation between the current pitch and the target pitch:
$$
r_{\text{pitch}} = 
\begin{cases} 
\exp \left( -4 (\theta_{\text{pitch}} - \theta_{\text{pitch}}^{\text{target}})^2 \right), & \text{if } d < d_{\text{th}}, \\
0, & \text{otherwise}.
\end{cases} \eqno{(3)}
$$


\subsubsection{Hollow Gap Kinematic Penalty ($r_{\text{hollow}}$)}
The most catastrophic failure mode on hollow stairs is kinematic locking, which occurs when a robot's foot slips into the empty space between two treads, severely trapping the leg. To strictly prohibit such behaviors, we design a hard penalty for any foot trajectory that intersects with the predefined hollow volumes as illustrated by the blue area in Figure~\ref{fig:foothold}(b). If the position of the $i$-th foot, denoted as $\mathbf{p}_{\text{foot}, i}$, enters the bounding box of the hollow gap region $B_{\text{hollow}}$,  a penalty of $- c_{\text{hollow}}$ is applied for that specific foot.  The reward is formulated as:
$$
r_{\text{hollow}} = \sum_{i=1}^{4} 
\begin{cases} 
- c_{\text{hollow}},  & \text{if } \mathbf{p}_{\text{foot}, i} \in B_{\text{hollow}}, \\ 
0, & \text{otherwise}.
\end{cases} \eqno{(4)}
$$
This penalty forces the robots to prioritize high and safe swing trajectories that completely clear the gaps.

    

\subsubsection{Stair Edge Penalty ($r_{\text{edge}}$)}
Even if a foot lands on the tread, stepping too close to the stair edge highly increases the risk of slipping off, especially given the noisy depth perception in the real world. As illustrated in the edge condition constraints, we introduce a penalty based on the distance $d_{\text{edge}, i}$ from the $i$-th foot's contact point to the stair edge depicted by the red dots in Figure~\ref{fig:foothold}(b). We define a safe margin threshold $d_{\text{safe}}$. The policy is penalized if the foot lands closer to the edge than this safety margin:
$$r_{\text{edge}} = \sum_{i \in \text{contact}} \begin{cases} -c_{\text{edge}},& \text{if } d_{\text{edge}, i} < d_{\text{safe}}, \\ 0, & \text{otherwise}. \end{cases} \eqno{(5)}$$
This encourages the robot to consistently target the center depth of the stair treads, maximizing stable footholds and significantly boosting sim-to-real robustness.

\subsection{Training Details}

\subsubsection{Terrain Curriculum Design}
To facilitate stable policy learning, we implement a terrain curriculum that progressively scales the terrains' complexity.

\begin{itemize}

\item Progressive Geometric Scaling: The curriculum initializes on flat ground ($0^\circ$) for basic locomotion learning. As performance improves, it progressively increases the step height while decreasing the tread depth and width, eventually reaching an extreme $55^\circ$ incline.

\item Structural Noise Injection: To enhance sim-to-real robustness, we inject random noise into vertical elevations and horizontal gap distances. This non-uniformity prevents gait memorization and forces reliance on real-time visuospatial perception.

\item Update and Resampling Mechanism: Robots advance upon reaching the top waypoint. To prevent catastrophic forgetting, masters of the highest level are reassigned to randomly sampled difficulties across the curriculum.

\end{itemize}

\subsubsection{Domain Randomization}
To facilitate sim-to-real transfer, we apply comprehensive domain randomization. This includes perturbing static and dynamic parameters (friction, mass, motor strength), applying external pushes and action delays, and randomizing the depth camera’s intrinsic FOV and mounting poses.

\section{EXPERIMENTS}

\subsection{Experimental Setup}
We train our policy in the Isaac Gym simulator using a single NVIDIA RTX 4090 GPU. For real-world deployment, we utilize a Unitree Go2 quadruped robot. To perceive the environment, an Intel RealSense D435 depth camera is mounted via a custom-designed bracket, capturing depth streams at 10 Hz. The trained policy is executed entirely onboard an NVIDIA Jetson Orin NX computing module. This module outputs target joint positions at a control frequency of 50 Hz, which are subsequently converted into motor torques by a low-level PD controller with stiffness and damping gains set to $K_p = 40$ and $K_d = 1$, respectively.

\subsection{Simulation Comparison and Ablation Study}
To evaluate the effectiveness of the StairMaster framework and the necessity of our proposed components, we conduct extensive comparative experiments in simulation. We compare our method against these baselines:
\begin{itemize}
    \item Extreme Parkour (EP) \cite{cheng2024extreme}: A two-stage visual parkour framework that distills a depth-based student policy from a privileged teacher.
    \item EP w/ Ours Rewards: EP trained with our hollow-stair-specific rewards, including $r_{\text{pitch}}$ and $r_{\text{foothold}}$.
    \item HIMLoco \cite{long2023hybrid}: A state-of-the-art blind locomotion policy based on a hybrid internal model.
    \item Ours w/o $r_{\text{pitch}}$: Our method without the waypoint-based pitch-tracking reward.
    \item Ours w/o $r_{\text{foothold}}$: Our method without the hollow-gap penalty $r_{\text{hollow}}$ and the stair-edge penalty $r_{\text{edge}}$.
    \item Ours w/o CA \& SRU: Our method with Cross-Attention replaced by direct feature concatenation and SRU replaced by a standard two-layer LSTM.
    \item Ours w/o SRU: Our method retaining Cross-Attention
    but replacing SRU with a standard two-layer LSTM.
    \item Ours w/o depth noise: Our method trained without sim-to-real depth-noise modeling.
\end{itemize}


We define two primary evaluation metrics: Success Rate and Average Reached Steps. We evaluate their performance across three distinct environments: a flat-ground baseline, standardized hollow stairs, and randomized mixed stairs. The mixed environment specifically consists of consecutive treads generated with stochastic step heights and random horizontal gaps between each pair of steps, posing a significant challenge to the robot's spatial reasoning and landing precision.

\begin{itemize}
    \item Success Rate: The average percentage of trials where the robot successfully reaches the final waypoint at the top platform on hollow stairs.
    
    \item Average Reached Steps: The average percentage of the steps successfully traversed per trial. A step is counted when the robot's center of mass passes the center of corresponding tread.
\end{itemize}
Each method was evaluated over 1000 episodes on each terrain. We report the results in Table \ref{tab:table_success_rate}. In comparison experiments, our method outperforms others in all metrics.

\begin{table}[t]
\renewcommand{\arraystretch}{1.15}
\centering
\setlength{\tabcolsep}{2.2pt}

\resizebox{\columnwidth}{!}{%
\begin{tabular}
{@{} l | c | c c c c c c | c c c c c c @{}}
\hline
\multirow{3}{*}{\textbf{Method}}
& \multicolumn{7}{c|}{\textbf{Success Rate (\%) $\uparrow$}}
& \multicolumn{6}{c}{\textbf{Average Reached Steps (\%) $\uparrow$}} \\
\cline{2-14}
& Flat
& \multicolumn{6}{c|}{Hollow Stairs}
& \multicolumn{6}{c}{Hollow Stairs} \\
& 0$^\circ$
& 20$^\circ$ & 30$^\circ$ & 40$^\circ$
& 50$^\circ$ & 55$^\circ$ & Mixed
& 20$^\circ$ & 30$^\circ$ & 40$^\circ$
& 50$^\circ$ & 55$^\circ$ & Mixed \\
\hline

\textbf{Ours}
& \textbf{100.0}
& \textbf{100.0} & \textbf{100.0} & \textbf{100.0}
& \textbf{98.0} & \textbf{97.5} & \textbf{86.5}
& \textbf{100.0} & \textbf{100.0} & \textbf{100.0}
& \textbf{99.1} & \textbf{98.4} & \textbf{93.0} \\

Ours w/o $r_{\text{pitch}}$
& 100.0
& 100.0 & 100.0 & 99.5
& 97.5 & 96.0 & 77.2
& 100.0 & 100.0 & 99.9
& 99.0 & 98.3 & 87.5 \\

Ours w/o $r_{\text{foothold}}$
& 100.0
& 100.0 & 100.0 & 100.0
& 97.5 & 97.0 & 83.5
& 100.0 & 100.0 & 100.0
& 98.6 & 98.2 & 91.2 \\

Ours w/o CA \& SRU
& 100.0 & 100.0 & 100.0 & 84.0 & 80.5 & 79.5 & 86.0
& 100.0 & 100.0 & 87.3 & 83.9 & 81.3 & 86.5 \\

Ours w/o SRU
& 100.0 & 100.0 & 100.0 & 87.5 & 81.0 & 81.0 & 83.5
& 100.0 & 100.0 & 89.0 & 83.4 & 89.4 & 84.7 \\

EP
& 100.0
& 0.0 & 0.0 & 0.0
& 0.0 & 0.0 & 0.0
& 17.0 & 19.4 & 0.0
& 0.0 & 0.0 & 0.5 \\

EP w/ Ours Rewards
& 100.0 & 0.0 & 0.0 & 0.0 & 0.0 & 0.0 & 0.0
& 21.4 & 20.2 & 0.0 & 0.0 & 0.0 & 1.1 \\

HIMLoco
& 100.0
& 0.0 & 0.0 & 0.0
& 0.0 & 0.0 & 0.0
& 2.9 & 2.9 & 0.0
& 0.0 & 0.0 & 0.0 \\

\hline
\end{tabular}%
}

\caption{Quantitative comparison results with other locomotion policies on hollow stairs in simulation.}
\label{tab:table_success_rate}
\end{table}

\begin{table}[h]
\renewcommand{\arraystretch}{1.2}
\centering
\setlength{\tabcolsep}{2.5pt}

\begin{tabularx}{\columnwidth}
{@{}>{\raggedright\arraybackslash}X | *{6}{c}@{}}
\hline
\multirow{3}{*}{\textbf{Method}}
& \multicolumn{6}{c}{\textbf{Average Collisions $\downarrow$}} \\
\cline{2-7}
& \multicolumn{6}{c}{Hollow Stairs} \\
& 20$^\circ$ & 30$^\circ$ & 40$^\circ$
& 50$^\circ$ & 55$^\circ$ & Mixed \\
\hline
\textbf{Ours}
& \textbf{0.03} & 0.04 & \textbf{0.78}
& \textbf{4.62} & \textbf{5.64} & \textbf{13.25} \\
Ours w/o $r_{\text{pitch}}$
& 0.06 & 0.16 & 1.59 & 9.68 & 12.83 & 31.13 \\
Ours w/o $r_{\text{foothold}}$
& 0.03 & \textbf{0.02} & 0.98 & 5.25 & 6.46 & 22.66 \\
\hline
\end{tabularx}

\caption{Comparison results of average collision counts in the
ablation study on hollow stairs.}
\label{tab:table_II}
\end{table}

\begin{figure*}[t] 
   \centering
   \includegraphics[width=\textwidth]{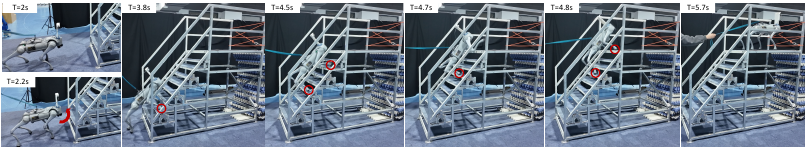}
   \caption{Snapshots of real-world experiments on the steep 55$^\circ$ hollow stairs.}
   \label{fig:real53}
\end{figure*}

\begin{figure}[t]
    \centering
    \includegraphics[width=0.48\columnwidth]{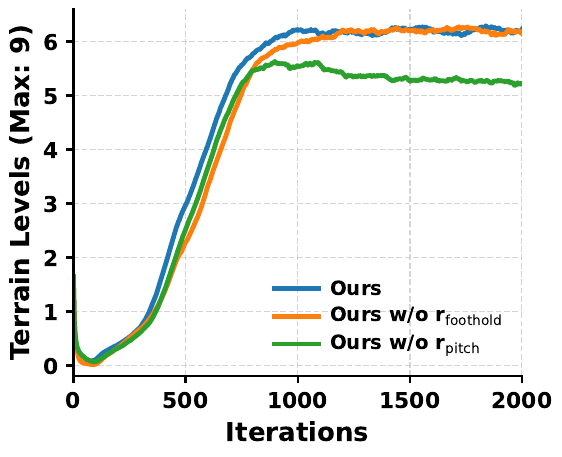}
    \hfill
    \includegraphics[width=0.48\columnwidth]{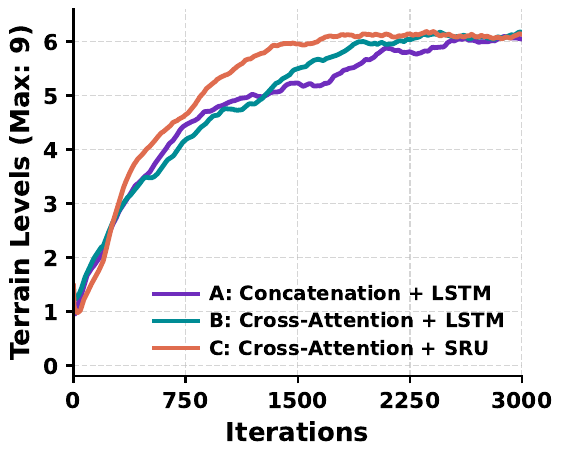}
    \caption{Terrain level comparison during training.}
    \label{fig:terrain_level_results}
\end{figure}

\begin{figure}[t] 
   \centering
   \includegraphics[width=0.98\columnwidth]{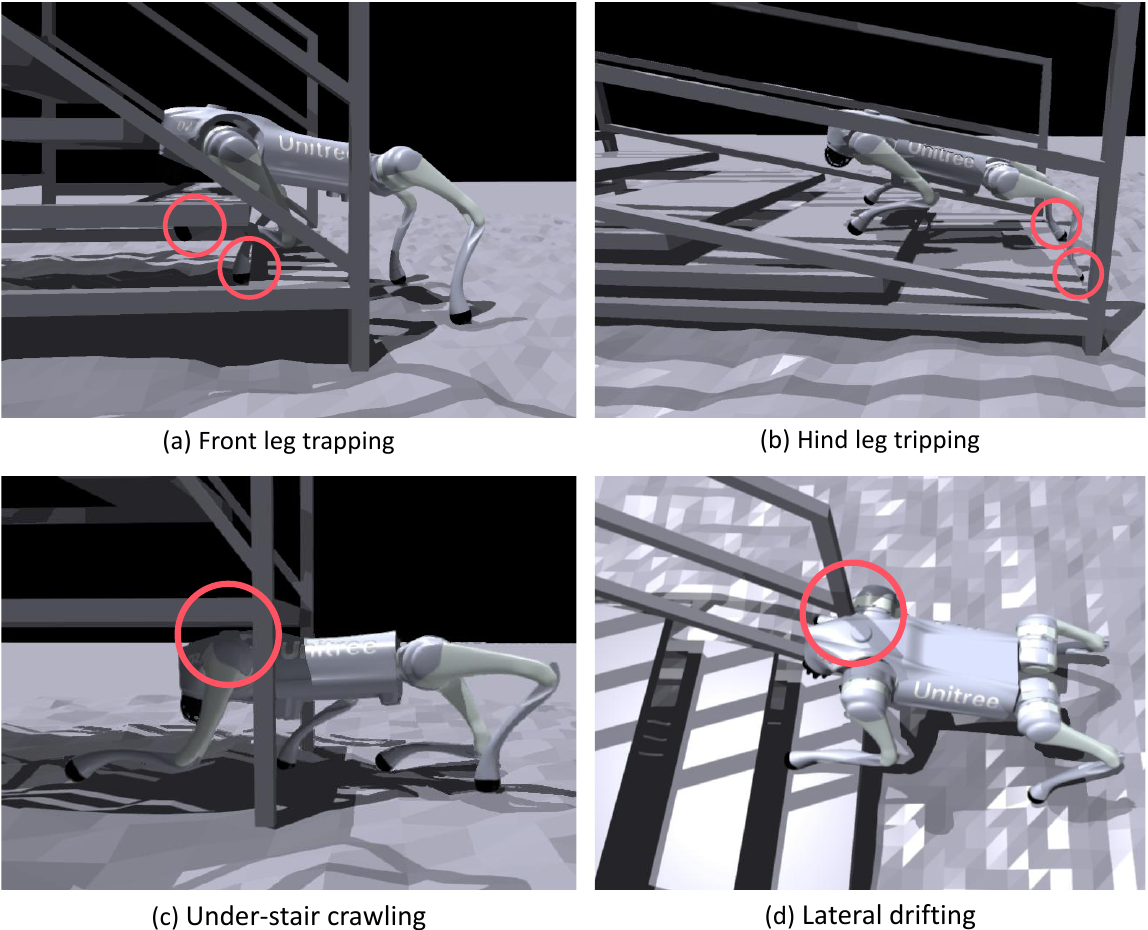}
   \caption{Classical failure cases in simulation.}
   \label{fig:failure}
\end{figure}

\textbf{1) Comparison with Baselines:}
As shown in Table~\ref{tab:table_success_rate}, both HIMLoco and Extreme Parkour achieve zero success on hollow stairs. Equipping Extreme Parkour with our hollow-stair rewards slightly improves its reached-step ratio at low inclines, but still results in zero successful trials. This indicates that reward adaptation alone is insufficient for handling the severe geometric discontinuities. In contrast, StairMaster achieves a $97.5\%$ success rate at $55^\circ$ and $86.5\%$ on randomized mixed stairs. The failure cases in Figure~\ref{fig:failure} further illustrate the common trapping, under-stair crawling, and lateral-drifting behaviors of the baselines.

\textbf{2) Effect of the Visuospatial Architecture:}
We compare Ours w/o CA \& SRU with Ours w/o SRU to isolate the effect of Cross-Attention, and compare Ours w/o SRU with the full model to evaluate SRU. Cross-Attention provides a modest improvement, while SRU substantially improves performance on steep and mixed stairs. As shown in Figure~\ref{fig:terrain_level_results}, the full architecture also progresses through the curriculum more rapidly, indicating that spatial-aware recurrence is the primary source of the architectural gain.

\textbf{3) Effect of Hollow-Stair Rewards:}
The reward ablations reveal complementary effects. Removing $r_{\text{pitch}}$ decreases mixed-stair success from $86.5\%$ to $77.2\%$ and increases mixed-terrain collisions from $13.25$ to $31.13$. Removing $r_{\text{foothold}}$ has a smaller effect on nominal success, but increases mixed-terrain collisions to $22.66$ and slows curriculum progression. Thus, $r_{\text{pitch}}$ mainly improves adaptation to steep and randomized stairs, while $r_{\text{foothold}}$ improves sample efficiency and foothold safety.

\subsection{Real-World Experiments}

To validate the sim-to-real transferability of the StairMaster framework, we conducted physical experiments on challenging hollow stairs with inclines of $37^\circ$ and $55^\circ$. Each policy was deployed on a Unitree Go2 quadruped and tested for 30 consecutive trials per stair. We also introduce the robot's built-in RL as an additional baseline for comparison. The results are shown in Table \ref{table_real_world}.

\begin{table}[h] 
\renewcommand{\arraystretch}{1.2} 
\centering

\newlength{\headerwidth}
\settowidth{\headerwidth}{\textbf{Hollow Stair Incline}}
\newcolumntype{E}{>{\centering\arraybackslash}p{0.5\headerwidth}}

\begin{tabular}{l | E E}
\hline 
\multirow{3}{*}{\textbf{Method}} & \multicolumn{2}{c}{\textbf{Success Rate (\%) $\uparrow$}} \\
\cline{2-3} 
 & \multicolumn{2}{c}{Hollow Stairs} \\
 & 37$^\circ$ & 55$^\circ$ \\
\hline 
\textbf{Ours}          & \textbf{93.33\%} & \textbf{63.33\%} \\
Ours w/o depth noise   & 50.00\% & 16.67\% \\
Extreme Parkour        & 0  & 0   \\
HIMLoco                & 0  & 0   \\
Built-in RL            & 46.67\% & 0   \\
\hline 
\end{tabular}
\caption{Real-world success rates on hollow stairs.}
\label{table_real_world}
\end{table}

As shown in Table~\ref{table_real_world}, StairMaster achieves success rates of $93.33\%$ and $63.33\%$ on the $37^\circ$ and $55^\circ$ hollow stairs, respectively, outperforming all evaluated baselines. Extreme Parkour and HIMLoco fail under both conditions. The Built-in RL policy reaches a $46.67\%$ success rate at $37^\circ$, but fails completely at $55^\circ$, indicating that proprioceptive locomotion alone is insufficient for the more severe geometric discontinuities of steep hollow stairs.

The necessity of our depth noise modeling is underscored by the Ours w/o depth noise ablation, which suffered a drastic success rate drop. The substantial performance drop indicates that depth noise modeling
plays an important role in sim-to-real transfer.

Our full framework demonstrates superior robustness and precision. As illustrated by the red circles in Figure~\ref{fig:real53}, StairMaster achieves highly accurate foot placement. By utilizing the visuospatial memory and $r_{\text{foothold}}$ supervision, the robot consistently lands in the center of the narrow treads, avoiding the trapping issues that plague the Built-in RL. On the challenging $55^\circ$ incline shown in Figure~\ref{fig:real53}, the effectiveness of $r_{\text{pitch}}$ is clearly visible. The robot proactively raises its pitch angle before reaching the first step ($T=2.2s$), allowing the camera to adapt to the steep environment and capture the upcoming stair structure. This look-ahead capability enables the robot to complete the $55^\circ$ climb in under 4 seconds, showcasing remarkable speed and fluid motion. Ultimately, real-world experiments confirm that our proposed rewards and architecture facilitate zero-shot sim-to-real transfer on discontinuous terrains.



\section{CONCLUSIONS}
In this work, we presented StairMaster, a robust reinforcement learning framework that enables quadruped robots to traverse high-risk hollow stairs. Our method effectively overcomes the challenges of sensor noise and visual occlusions by combining a Cross-Attention-based visuospatial encoder with a Spatial-Aware LSTM. We introduced a three-stage training pipeline and tailored reward functions for active pitch control and precise foothold placement to ensure the robot's safety and stability on extremely sparse terrains. Real-world experiments on the Unitree Go2 robot confirm that our policy achieves remarkable robustness, successfully navigating 55° hollow stairs with zero-shot sim-to-real transfer.

Future work will focus on improving sim-to-real fidelity through motor modeling and integrating RGB images to further enhance environmental awareness. We also aim to extend our framework to handle more complex industrial constraints, pushing the boundaries of robot mobility in extreme scenarios.

\bibliography{aaai2027}


\end{document}